\documentclass[letterpaper, 10 pt, conference]{ieeeconf}  

\IEEEoverridecommandlockouts                              
\usepackage{graphicx} 
\graphicspath{{figures/}}

\usepackage{color}
\usepackage{latexsym}
\usepackage{multicol}

\usepackage{graphicx} 
\usepackage[font=small,labelfont=bf]{caption}
\usepackage{lipsum}
 \usepackage{subcaption}

\newtheorem{definition}{Definition}
\newtheorem{problem}{Problem}
\newcommand*{\mydprime}{^{\prime\prime}\mkern-1.2mu}

\usepackage{amsmath} 

\usepackage{algpseudocode}
\usepackage{algorithm}
\usepackage[subnum]{cases}



\renewcommand{\thefigure}{\arabic{figure} }


\title{\LARGE \bf
A Hierarchical Reinforcement Learning Method for Persistent Time-Sensitive Tasks
}

\author{Xiao Li and Calin Belta
\thanks{X. Li is with the Department of Mechanical Engineering. C. Belta is with the Division of Systems Engineering and the Department of Mechanical Engineering at Boston University, Boston, MA -2215. Email: \{xli87,cbelta\}@bu.edu}
\thanks{This work is partially supported by the ONR under grant N00014-14-1-0554 and by the NSF under grant NRI-1426907, CMMI-1400167}
}

\begin{document}

\maketitle
\thispagestyle{empty}
\pagestyle{empty}

\begin{abstract}

Reinforcement learning has been applied to many interesting problems such as the famous TD-gammon \cite{TDgammon} and the inverted helicopter flight \cite{ng}. However little effort has been put into developing methods to learn policies for complex persistent tasks and tasks that are time-sensitive. In this paper we take a step towards solving this problem by using signal temporal logic (STL) as task specification, and taking advantage of the temporal abstraction feature that the options framework provide. We show via simulation that a relatively easy to implement algorithm that combines STL and options can learn a satisfactory policy with a small number of training cases.

\end{abstract}

\section{INTRODUCTION}

Reinforcement learning is the problem of learning from interaction with the environment to achieve a goal \cite{introRL}. Usually the interaction model is unknown to the learning agent and an optimal policy is to be learned with sequences of interaction experiences and a reward that indicates the "correctness" of taking an action hence the reinforcement. There has been a number of successful attempts to apply reinforcement learning to the field of control. One of the most widely known efforts is learning of a flight controller for aggressive aerobatic maneuvers on a RC helicopter \cite{ng}. In addition, a PR2 (Personal Robot 2) has learned to perform a number of household chores such at placing a coat hanger, twisting open bottle caps, etc using ideas from reinforcement learning \cite{Levine2015}. More recent efforts in this area has led a learning agent to play many of the classic Atari 2600 games at the professional human level \cite{Mnih2015}, and the the possibility of a match at the game of Go between AlphaGo (an AI agent created by Google Deepmind \cite{Silver2016}) and one of the top Go players in the world Lee Sedol.

Reinforcement learning has great potential in areas where precise modeling of system and environmental dynamics are difficult but interaction data is available, which is the case for many real world applications. In classical reinforcement learning, the reward structure needs to be carefully designed to obtain a desirable outcome, and often additional techniques such as reward shaping \cite{Ng1999},\cite{shaping} need to be applied to improve the learning efficiency. Moreover, the tasks being learned are often single goal episodic tasks such as reaching a destination in shortest time \cite{Dutech2005}, paddling a ball \cite{Kober2010} or winning a game that has a set of well defined rules \cite{Mnih2015},\cite{TDgammon}. Little effort has been put into creating a learning agent for complex \textit{time-sensitive multi-goal persistent tasks}. Persistence requires that the task is  continuous/cyclic and does not have a notion of termination (or absorbing state), whereas multi-goal time-sensitiveness indicates that the task consists of subtasks and it is desirable to switch among them in a predefined timely fashion. An example of such a task is controlling of a robotic manipulator on an assembly line. Here the manipulator may switch from fastening a screw at one location to wielding at another location, and the time between the switch may need to be controlled depending on how the position and orientation of the part are handled by possibly the conveyer belt or other manipulators. 

Learning of simple persistent tasks has traditionally been tackled using average reward reinforcement learning \cite{Mahadevan1996}. Well known algorithms include R-learning \cite{Schwartz1993} and H-learning \cite{Hlearning}. However these methods work well with only unichain MDPs where every deterministic and stationary policy on
 that MDP contains only a single loop, also called a recurrent class of states \cite{Puterman1994}. This is obviously not enough for any task of reasonable complexity. \cite{Fu2014},\cite{Sadigh2014} uses model-based approaches to learn policies that maximize the probability of satisfying a given linear temporal logic (LTL) formula. However, using probability of satisfaction to guide learning can be of low efficiency because no "partial credit" is given to the agent for being "close" to satisfying the specification. And thus the agent performs random search before it "accidentally" satisfies the LTL specification for the first time. Moreover, LTL has time-abstract semantics that prevent users from specifying time bounds quantitatively. 
 
In this paper we turned to signal temporal logic (STL), a rich predicate logic that can describe tasks involving bounds on physical states, continuous time windows and logical relationships. For example the assembly line manipulator control task described earlier can be easily expressed using STL in the form "from start of the assembly task until the end, with a period of $\Delta t$ repeatedly position the end-effector to within a tolerance of the screw location and perform fastening motion, and then position the end-effector to within a tolerance of the wielding point and perform the wielding task" (the STL formula is presented in the next section). The one significant convenience that STL brings is its equipment with a continuous measure of satisfiability called the \textit{robustness degree}, which translates naturally to a continuous reward structure that can be used in the reinforcement learning framework. Therefore with STL the user only has to "spell out" the task requirements in a compact yet powerful language and the rest will be taken care of (no need to struggle with designing a good reward structure).
 
The challenge of using STL is that evaluation of the robustness degree requires a state trajectory, therefore either some kind of memory needs to be incorporated into the learning agent or the state/action space be expanded to incorporate trajectories that the agent can choose from. Here we adopt the options framework \cite{between} which abstracts each subtask as an MDP with a policy of its own, and a higher level policy is present to choose among the subtask policies at appropriate times. In this paper we present an algorithm that given an STL task specification, automatically generates a set of subtasks, and by interacting with the environment simultaneously learn the subtasks' policies and the higher level policy that will lead the agent to satisfy the given specification.

Section \ref{sec:2} introduces the Q-learning algorithm that subsequent contents are developed on, as well as the options framework and STL. Section \ref{sec:3} describes in detail the proposed algorithm. Section \ref{sec:4} provides simulation results to verify the proposed approach and some discussions about the advantages and shortcomings of the algorithm. Section \ref{sec:5} concludes with final remarks and directions for future work.

\section{BACKGROUND}
\label{sec:2}
Reinforcement learning bears the curse of dimensionality. Especially for discretized representations of state and action spaces (used in many classical tabular methods
 \cite{introRL}), the number of parameters (state value, action value, etc) increase exponentially with the size of the state/action space. One attempt to alleviate such computational burden is to exploit temporal and state abstractions and the possibility of learning on a reduced set of abstractions as oppose to the primitive state and actions. Ideas along 
 this line are called hierarchical reinforcement learning (HRL) in the literature, and a survey of advances in this area as well as the main approaches used are provided in 
 \cite{Barto2003}.  We base our work on the options framework developed in \cite{between} for its ability to deal with temporally extended actions, which is an 
 extremely helpful factor in the development of an algorithm that learns a policy that satisfies the complicated task specification given by an STL formula.

\subsection{Reinforcement Learning Framework and Q-Learning}
\label{subsec:2.1}
Here we briefly describe the reinforcement learning framework for discrete-time finite Markov decision processes (MDP). 

\begin{definition}
\label{def:1}
An MDP is a tuple $\langle S,A,T(\cdot,\cdot,\cdot),R(\cdot,\cdot,\cdot)\rangle$ where 

\begin{itemize}
\item $S$ is a finite set of states;
\item $A$ is a finite set of actions;
\item $T: S \times A \times S \rightarrow [0,1]$ is the transition probability with $T(s,a,s^{\prime})$ being the probability of taking action $a \in A$ at state $s \in S$ and end up in state $s^{\prime} \in S$;
\item $R: S \times A \times S \rightarrow {\rm I\!R}$ is the reward function with $R(s,a,s^{\prime})$ being the reward obtained by taking action $a$ in $s$ and end in $s^{\prime}$.
\end{itemize}

\end{definition}
\noindent In reinforcement learning, the transition model $T(s,a,s^{\prime})$ and the reward structure $R(s,a,s^{\prime})$ are unknown to the learning agent (but an immediate reward $r$ is given to the agent after each action), instead the agent has to interact with the environment and figure out the optimal sequence of actions to take in order to
 maximize the obtained reward. We have based our method on one of the most popular model-free off-policy reinforcement learning algorithms called Q-learning \cite{Watkins}. In short, Q-learning aims at finding a policy $\pi: S \rightarrow A$ that maximizes the expected sum of discounted reward given by 
 
 \begin{equation}
 \label{eq:0}
 V(s)=E[\sum_{i=0}^{\infty} \gamma^i r_i].
 \end{equation}
 
 \noindent Here $\gamma \in$ [0,1] is a constant discount factor and is decayed with time (hence the exponent $i$) to put higher value on more recent rewards. $r_i$ is the one step immediate reward at step $i$. Equation (\ref{eq:0}) can be written recursively as 
 
 \begin{equation}
 \label{eq:-1}
V(s) = \sum_{s^{\prime} \in S} T(s,a,s^{\prime})(r + \gamma V(s^{\prime})),
 \end{equation}
 
 \noindent which becomes the well known Bellman's Equation. Algorithms exist that learns the optimal value function $V(s)$ from experience. The most famous one is perhaps the temporal difference learning algorithm (also called TD-learning \cite{TD}). After $V(s)$ converges to its optimal value $V^{\star}(s)$, we have the recursive relationship
 
 \begin{equation}
 \label{eq:-3}
V^\star(s) = \underset{a \in A}{\max}\sum_{s^{\prime} \in S} T(s,a,s^{\prime})(r + \gamma V^{\star}(s^{\prime})).
\end{equation}

 \noindent And the optimal policy is calculated from 
 
\begin{equation}
\label{eq:-4}
\pi^{\star}(s) = \underset{a \in A}{\arg\max} V^{\star}(s). 
\end{equation}

\noindent However without knowing the transition model $T(s,a,s^{\prime})$, it is difficult to extract the optimal policy from $V^\star(s)$. This is where Q-learning comes in. Define an action-value function that assigns a value to each state-action pair (also known as the Q function) as follows
 
 \begin{equation}
 \label{eq:-2}
 Q(s,a) = \sum_{s^{\prime}} T(s,a,s^{\prime})( r+ \gamma V(s^{\prime})).
 \end{equation}
 
\noindent Then following Equation (\ref{eq:-3}) we have 

\begin{equation}
\label{eq:-5}
V^\star(s) = \underset{a \in A}{\max}Q^\star(s,a).
\end{equation}

\noindent Now we can write the optimal Q-function in a recursive form by

\begin{equation}
 Q^\star(s,a) = \sum_{s^{\prime} \in S} T(s,a,s^{\prime})( r+ \gamma  \underset{a \in A}{\max}Q^\star(s,a)).
\end{equation}

\noindent And $Q^\star(s,a)$ can be approximated by calculating a running average of the Q-values obtained from experience.

Assume at time $t$ the agent takes action $a$, transitions from state $s$ to $s^{\prime}$, and obtains a one step immediate reward $r$ (experiences usually take form of a tuple $(s,a,s^{\prime},r)$). The Q-function is then update following 
 
 \begin{equation}
 \label{eq:1}
 Q(s,a) \leftarrow Q(s,a) + \alpha_t(r+\gamma \underset{a^{\prime} \in A}{\max}Q(s^\prime,a^\prime) - Q(s,a)),
 \end{equation}

\noindent where $\alpha \in [0,1]$ is the learning rate.  It is proven in \cite{convergence} that if the choice of $\alpha$ satisfies $\sum_t^{\infty}\alpha_t = \infty$ and $\sum_t^{\infty}\alpha^2_t < \infty$ while every state and action are visited infinitely often, then $Q(s,a)$ converges (denoted by $Q^{\star}(s,a)$). In practice it is usually sufficient to use a constant $\alpha$ and thus the subscript $t$ is dropped in later formulations. After convergence, the optimal policy can be calculated by

\begin{equation}
\label{eq:8}
\pi^{\star}(s) = \underset{a \in A}{\arg\max} Q^{\star}(s,a).
\end{equation}

\noindent Since action $a$ is an explicit variable of the Q-function, Equation (\ref{eq:8}) can be easy evaluated.

\subsection{Option-Based Hierarchical Reinforcement Learning}

The options framework deals explicitly with temporally extended actions. An option is defined by a tuple $\langle I, \pi,\beta \rangle$ where $I \subseteq S$ is the initiation set denoting the states where an option is available. $\pi:S \rightarrow A$ is the option's policy (also called a flat policy) and $\beta: S \rightarrow [0,1]$ is the termination map defining the probability of termination of an option at each state. Suppose at time $t$ the agent resides at state $s$. Instead of choosing an action $a \in A$, the agent chooses an option
$o = \langle I_o,\pi_o,\beta_o\rangle \in O$, where $O$ is the set of options (note that option $o$ needs to be available at state $s$ i.e. $s \in I_o$). After selecting the option, the agent follows the option's flat policy $\pi_o(s)$ until termination is invoked. If the termination state is $s^{\prime}$, then $\beta(s^{\prime}) > 0$. Analogous to Q-learning, the experience that the agent obtained now becomes a tuple $(s,o,s^{\prime},r_o)$, where $r_o$ is a lumped reward from executing option $o$ to termination. Assuming that option $o$ is being executed for $k$ time steps, now instead of updating an action-value function $Q(s,a)$, an option-value function is updated using 

\begin{equation}
\label{eq:3}
Q(s,o) \leftarrow Q(s,o) + \alpha(r_o+\gamma^k \underset{o^\prime \in O}{\max}Q(s^\prime,o^\prime) - Q(s,o)).
\end{equation}

\noindent This update is applied each time an option is executed to termination. Equation (\ref{eq:3}) is very similar to Equation (\ref{eq:1}) except for the exponent $k$ on the discount factor $\gamma$. This is to signify that the option is executed for a temporally extend period of time and future rewards should be discounted accordingly. 
It is worth mentioning that a primitive action can be considered a one step option $o_a = \langle I_a,\pi_a,\beta_a\rangle$ where $I_a = S$, $\pi_a(s) = a$ and $\beta_a(s) = 1,\forall s \in S$, therefore if $o_a$ is executed at all times then Equation (\ref{eq:3}) becomes Equation (\ref{eq:1}). The optimal options policy $\mu:S \rightarrow O$ is obtained by 

\begin{equation}
\label{eq:4}
\mu^{\star}(s) = \underset{o \in O}{\arg\max} Q^{\star}(s,o).
\end{equation}

\noindent The flat policy $\pi_o(s)$ for each option in $O$ can be provided by the user or be learned simultaneously with the options policy $\mu(s)$. Details on simultaneous learning will be discussed in the next section. We refer readers to \cite{between} for a detailed formulation of the options framework.

\subsection{Signal Temporal Logic (STL)}

Signal temporal logic is a framework used to describe an expressive collection of specifications in a compact form. It was originally developed to monitor continuous-time 
signals, but can be extended to describe desired state constraints in a control system. Here we briefly present the necessary definitions of STL and refer interested readers 
to \cite{STL1}, \cite{STL2}, \cite{SadraSTL} for further details. Informally, STL formulas consist of boolean connectives $\neg$(negation/not), $\wedge$(conjunction/and), $
\vee$ (disjunction/or), as well as bounded-time temporal operators $U_{[t_1,t_2)}$ (until between $t_1$ and $t_2$), $\diamondsuit_{[t_1,t_2)}$ (eventually between $t_1$ and $t_2$) and  $\Box_{[t_1,t_2)}$ (always between $t_1$ and $t_2$) that operate on a finite set of predicates over the underlying states. As a quick example, consider a robot traveling in a plane with its position $(x,y)$ as states. The trajectory of the robot is specified by a simple STL formula 

\begin{equation}
\label{eq:5}
\begin{split}
\phi &= \Box_{[0,4)}\psi,\\
\psi &=\Box_{[0,4)}[(x > 10)\wedge(x < 14)\wedge(y > 6)\wedge(y < 10)].
\end{split}
\end{equation}

\noindent The formula in Equation (\ref{eq:5}) reads "always in 0 to 4 time steps, x is to be greater than 10 and smaller than 12, and y greater than 6 and smaller than 8", which specifies that the robot should stay in a square region given by bound $(x_{max}=14,x_{min}=10,y_{max}=10,y_{min}=6)$ from 0 to 4 time steps.

In this paper, we constrain STL to be defined over sequences of discrete valued states produced by the MDP in Definition (\ref{def:1}). We denote $s_t$ to be the state at time $t$, and $s_{t:t+k}$ to be a time series of the state trajectory from $t$ to $t+k$, i.e. $s_{t:t+k} = [s_{t},s_{t+1}, \dots ,s_{t+k}]$. The usefulness of STL lies in its equipment with a set of quantitative measure of how well a given formula is satisfied, which is called \textit{robustness degree} (robustness for short). In the above example, a term like $x>10$ is called a predicate which we denote by \textit{p}. Let \textit{p} take the form of a general inequality $f(s) < c$, where $f(s)$ is a function of the states and $c$ is a constant (such as $f(x,y) = x+2y < 7$). If a state trajectory $s_{t:t+k}$ is provided, the robustness of an STL formula is defined recursively by 

\begin{equation}
\label{eq:6}
\begin{split}
&r(s_t,f(s_t)<c) = c - f(s_t),\\
&r(s_t,\neg \phi) = -r(s_t,\phi),\\
&r(s_t,\phi_1\wedge \phi_2) = \min(r(s_t,\phi_1),r(s_t,\phi_2)),\\
&r(s_t,\phi_1 \vee \phi_2) = \max(r(s_t,\phi_1),r(s_t,\phi_2)),\\
&r(s_{t:t+k},\Box_{[t,t+k]}\phi) = \underset{t^{\prime} \in [t,t+k)}{\min}(r(s_{t^{\prime}},\phi)),\\
&r(s_{t:t+k},\diamondsuit_{[t,t+k]}\phi) = \underset{t^{\prime} \in [t,t+k)}{\max}(r(s_{t^{\prime}},\phi)),\\
&r(s_{t:t+k},\phi_1 U_{[t,t+k)}\phi_2) =\\
& \underset{t^{\prime} \in [t,t+k)}{\max}\left( \min \left(r(s_{t^{\prime}},\phi_1), \underset{t^{\mydprime} \in [t,t^{\prime})}{\min}r(s_{t\mydprime},\phi_2)\right)\right ).
\end{split}
\end{equation}

\noindent Note that in general if formula $\phi$ contains temporal operators ($\Box, \diamondsuit, U$), a state trajectory $s_{t:t+k}$ is required to evaluate robustness, but if $\phi$ contains only boolean connected predicates, the robustness is then evaluated with respective to one particular state $s_t$. Using the above definition of robustness, a larger positive value means stronger satisfaction and a larger negative value means stronger violation of the STL formula.

\begin{table}[]
\centering
\caption{Simple STL Example}
\label{tab:1}
\begin{tabular}{|c|c|c|c|c|}
\hline
\textbf{$t$}              & 0     & 1      & 2      & 3      \\ \hline
\textbf{$s_t=(x_t,y_t)$}  & (9,7) & (10,7) & (11,7) & (11,8) \\ \hline
\textbf{$r(s_t,\psi)$}           & -1    & 0      & 1      & 1      \\ \hline
\textbf{$r(s_{0:4},\phi)$} & \multicolumn{4}{c|}{-1}          \\ \hline
\end{tabular}
\end{table}

Table \ref{tab:1} shows an example of how to calculate the robustness of a trajectory given STL formula in Equation (\ref{eq:5}). We can see that for this trajectory the overall robustness is negative meaning that Equation (\ref{eq:5}) is violated. The reason is that the first point in the trajectory lies outside the desired square given by $\psi$ and the STL formula dictates that all positions should stay inside the square within the timeframe of 0 to 4 . If instead of $\Box_{[0,4)}\psi$ we specify $\diamondsuit_{[0,4)}\psi$, then $r(s_{0:4},\phi) = 1$ because $\diamondsuit$ (eventually) looks at the highest satisfying point whereas $\Box$(always) looks at the highest violation point. The point of maximum satisfaction occurs at the center of the square ($s = (12,8)$) with a robustness value of 2.

Even though the example above uses the simplest form of STL formula for explanation, an STL specification can be much richer. For the assembly line manipulator task mentioned in the Introduction, let $p_{ee}$ be the position of the end-effector, $p_{screw}$ be the position of the screw to be fastened, and $p_{wield}$ be the wielding point. Then the assembly task can be expressed by the STL formula

\begin{equation}
\label{eq:7}
\begin{split}
\phi_{assembly} = \Box_{[0,\infty)}[&\diamondsuit_{[0,\Delta t)}(|p_{ee}-p_{screw}|<\eta_{screw}) \wedge\\
& \diamondsuit_{[0,\Delta t)}(|p_{ee}-p_{wield}|<\eta_{wield})].
\end{split}
\end{equation}

\noindent In the above formula $|\cdot|$ is the Euclidean distance. $\eta_{screw}$ and $\eta_{wield}$ are the position thresholds for the screw fastening and wielding tasks respectively.

\section{REINFORCEMENT LEARNING FOR STL SPECIFIED GOALS}
\label{sec:3}

The options framework provides a way to expand the action space to a set of options. Executing options generate repeatable trajectories that can be used to evaluate STL robustness. In this section we present an algorithm that, given an STL formula that describes 
the desired behavior of the system, automatically generates a set of options. The algorithm then learns a hierarchically optimal options policy and all options' flat policies by interacting with the environment (more on hierarchical optimality in the next section).
\subsection{Problem Formulation}

Given an options policy $\mu:S \rightarrow O$, let $V^\mu(s)$ be the expected sum of discounted lumped reward of state $s$ obtained from following $\mu$, which can be written recursively as 

\begin{equation}
\label{eq:8}
V^\mu(s) = E[r_o + \gamma_o^{k_o}V^\mu(s^\prime)].
\end{equation}

\noindent In the above equation, the subscripts $o$ denote the option being executed i.e. $o=\mu(s)$ at each state $s$. $r_{o} = r(s_{t:t+k_o}, \phi)$ is the lumped reward obtained from executing option $o$ at time $t$ and state $s$, and terminating at time $t+k_{o}$ and state $s^\prime$ (refer to Equation (\ref{eq:6}) for notation and robustness calculation). Here we denote $k_{o}$ to be the number of time steps option $o$ takes to terminate. The problem that we address in this paper can then be formulated as:

\vspace{2mm}
\begin{problem}
\label{prob:1}
Given an MDP $M = \langle S,A,T,R \rangle$ with unknown transition model $T(s,a,s^{\prime})$ and reward structure $R(s,a,s^\prime)$, an STL formula $\phi$ over $S$, and a set of options $O$, find a policy $\mu:S \rightarrow O$ that maximizes the expected sum of discounted lumped reward as specified in Equation (\ref{eq:8}). 
\end{problem} 
\vspace{2mm}


Before the algorithm is presented, we introduce some terminology. First a \textit{primitive option} is an option whose policy 
is a flat policy (i.e $\pi_{p}: S \rightarrow A$). This is in contrast with a \textit{hierarchical option} whose policy 
maps states to lower level options ($\pi_{h}: S \rightarrow O$). In other words a hierarchical option is an option over option and 
thus higher up the hierarchy. We will not be using hierarchical options in this paper. A \textit{temporally combined option} is an 
option constructed from executing a selected set of options in a predefined order. For example, suppose we have two 
primitive options $o_{p1} = \langle I_{p1}, \pi_{p1},\beta_{p1} \rangle$ and $o_{p2} = \langle I_{p2}, \pi_{p2},\beta_{p2} \rangle$, 
a temporally combined option $o_{p1-2} = \langle I_{p1-2}, \pi_{p1-2}, \beta_{p1-2} \rangle$ can be executed by first following option $o_{p1}$ until termination and then follow option $o_{p2}$ until termination. 
Therefore the initiation set $I_{p1-2} = I_{p1}$ and the termination map $\beta_{p1-2} = \beta_{p2}$. Also it should be ensured 
that the states where termination of option $o_{p1}$ is possible should be an element of the initiation set of $o_{p2}$ i.e. 
$\{s:\beta_{p1}(s) > 0\} \subseteq I_{p2}$. A temporally combined option can be a primitive option or a hierarchical option depending on its 
constituent options. For the method presented in this paper, all options are primitive options hence the subscript $p$ is dropped.

\subsection{The Hierarchical STL Learning Algorithm (HSTL-Learning)}

Given a STL specification containing $n$ boolean connected predicates $\psi_1$ to $\psi_n$ (like the $\psi$ in Equation 
(\ref{eq:5})), for each $\psi_i \in \{ \psi_1, \dots, \psi_n \}$, construct a primitive option $o_i = \langle I_i, \pi_i, \beta_i \rangle$ (
$I_i$ and $\beta_i$ are user defined). Using these primitive options, a set of temporally combined options $O_T$ is constructed. The way in which $O_T$ is constructed can be controlled by the user.  For example if 
the primitive options set is $O_p = \{o_1,o_2,o_3\}$ , a possible temporally combined options set can be $O_T = \{o_1,o_2,o_3,o_{1-2},o_{1-3},o_{2-3}, o_{1-2-3}\}$. Here we take advantage of the fact that Q-learning is an off-policy learning algorithm, meaning that the learned policy is independent of the exploration scheme \cite{offpolicy}. Hence multiple policies can be learned simultaneously while the agent is interacting with the environment. In the case of this example, $n+1$ policies need to be learned where $n$ is the number of boolean connected predicates in the STL specification (hence the number of flat policies) and one more for the options policy $\mu: S \rightarrow O_T$. The complete learning algorithm is present in Algorithm \ref{alg:1}.

\begin{algorithm}
\caption{HSTL-Learning}
\label{alg:1}
\begin{algorithmic}[1]
\Procedure{HSTL-update}{$\phi,o,Traj, Act$}
\State For each of the $n$ primitive options, initialized action-value function $Q(s,a) \leftarrow Q_0(s,a)$, initiation set $I$ and termination map $\beta$
\State Construct the temporally combined options set $O_T$
\State Initialize the option-value function $Q_i(s,o) \leftarrow Q_0(s,o)$ for $o \in O_T$
\State Choose learning rates $\alpha$ and discount factors $\gamma$ for all learning agents
\State $s_k \leftarrow Traj[:,k]$ \Comment this is the state where $o$ is terminated. colon indicates all elements in the dimension
\For{$i=1$ to $k-1$ }
\State $s_i \leftarrow Traj[:,i]$
\State $a_i \leftarrow Act[:,i]$
\State $s_{i+1} \leftarrow Traj[:,i+1]$

\For{$j=1$ to $n$}  \Comment  update all primitive options' Q-functions
\State $r_j \leftarrow r(s_{i+1},\psi_i)$ \Comment robustness as the reward for flat policy learning, refer to Equation (\ref{eq:6})
\State $ Q_j(s_i,a_i) \leftarrow Q_j(s_i,a_i) + \alpha_j(r_j+\gamma_j \underset{a_{i+1} \in A}{\max}Q_j(s_{i+1},a_{i+1}) - Q_j(s_i,a_i))$
\EndFor

\State $Traj_{seg}=Traj[:,i:k]$ \Comment $i:k$ indicates element $i$ to $k$
\State $r_{o_i} = r(Traj_{seg},\phi)$
\State $Q(s_i,o) \leftarrow Q(s_i,o) + \alpha_o(r_{o_i}+\gamma_o^k \underset{o_k \in O_T}{\max}Q(s_k,o_k) - Q(s_i,o))$
\EndFor

\Return all $Q_j(s,a)$ for $j \in 1,\dots,n$ and $Q(s,o)$
\EndProcedure

\end{algorithmic}
\end{algorithm}

The inputs to Algorithm \ref{alg:1} are an STL specification $\phi$, the currently selected option $o \in O_T$, and the trajectory resulted from executing $o$ to termination $Traj$. Here $Traj$ is a $m \times k$ matrix where $m$ is the dimension of state space and $k$ is the number of time steps $o$ is executed before termination. $Act$ is a $q \times k$ matrix where $q$ is the dimension of primitive action space. The algorithm outputs the updated $n+1$ Q-functions. The main idea of Algorithm \ref{alg:1} is that every time an option is executed to termination, the resulting trajectory is used to calculate a reward based on evaluating its robustness against the given STL formula (line 19). This reward is used to update the Q-function $Q(s,o)$. In cases where the time of executing an option to termination is less than that required to evaluate the robustness of the given STL formula, the upper time bound of the STL formula is adjusted to coincide with the execution time of the option, and evaluation is proceeded as usual. This is to ensure that choices of options are Markovian and does not depend on previous history. In addition, every primitive step $(s,a,s^{\prime})$ within the trajectory is used to update the Q-function $Q(s,a)$ for all options' flat policies, with the reward being the robustness of the resulting state $s^{\prime}$ with respective to the corresponding $\psi$ (line 15). Because $Q(s,o)$ is updated once only when an option terminates, convergence to a desirable policy can be quite slow. To speed up the learning process, an \textit{intra-option update step} is introduced which follows from the idea of intra-option value learning presented in \cite{between}. If an option is initiated at state $s_t$ and terminated at $s_{t+k}$ with trajectory $s_{t:t+k}$, then for every intermediate state $s_i, i \in [t,t+k]$ we can also consider the sub-trajectory $s_{i:t+k}$ a valid experience, where option $o$ is initiated at state $s_i$ and terminated at $s_{t:t+k}$. Therefore instead of updating $Q(s,o)$ only once for state $s_t$, it is updated for all intermediate states (lines 18-20), which drastically increases the efficiency for experience usage.

\subsection{Discussion} 
\label{subsec:3.3}

In this subsection we discuss some of the advantages and shortcomings of the proposed method. Unlike conventional reinforcement learning approaches where manual design of rewards is necessary, STL provides a way to conveniently specify complicated task goals while naturally translates the specifications to rewards. In addition, since robustness is a continuous measure of satisfiability, the resulting reward structure helps to speed up learning of the flat policies much like potential-based reward shaping \cite{shaping}. 

The correctness and completeness of the proposed algorithm are determined by the options framework. Here we introduce the notion of hierarchical optimality. A policy is said to be hierarchically optimal if it achieves the highest cumulative reward among all policies consistent with the given hierarchy \cite{Dietterich2000}. In general, a hierarchical learning algorithm with a fixed set of options converges to a hierarchically optimal policy \cite{between}, which is the case for the HSTL-learning algorithm. More specifically, the HSTL-learning algorithm will find a hierarchically optimal policy $\mu^\star$ that satisfies

\begin{equation}
\mu^\star(s) = \underset{\mu}{\arg\max}V^\mu(s)
\end{equation}

\noindent for a fixed set of options ($V^\pi$ defined in Equation (\ref{eq:8})). Whether robustness of the STL specification is satisfied/maximized depends on the set of options provided to the algorithm. A policy leading to trajectories that maximize the robustness of the given STL formula will be found if the trajectories can be constructed from the options provided. Therefore the correctness and completeness of the proposed algorithm are related to the hierarchical optimality property, and hence also depend on the set of options provided.

On complexity, Algorithm \ref{alg:1} requires $k+n$ operations per update. Here $k$ is the number of steps the current option takes to terminate, and $n$ is the number of elements in the set $O_T$. $n$ depends on the number of flat policies and how $O_T$ is constructed. Like Q-learning, the number of training steps required for convergence depends largely on the learning parameters listed in Table \ref{tab:2}, and convergence is guaranteed if each state-action pair is visited infinitely often (convergence guarantee discussed in Section \ref{subsec:2.1}).

Finally, it is worth mentioning that multiple trajectories exist that maximally satisfy a given STL formula (for example any trajectory that passes through $x = 1$ maximally satisfy $\phi = \diamondsuit_{[0,t)}[(x>0)\wedge (x < 2)]$). The proposed method chooses only the most greedy trajectory given the set of available options. This takes away some flexibility and the diversity of policies an agent can learn, but is also a predictable characteristic that can be used towards one's advantage. 

\section{CASE STUDY}
\label{sec:4}
In this section we evaluate the performance of the proposed method in a simulated environment, and provide a discussion of the results. As depicted in Figure (1), a mobile robot navigates in a $15 \times 15$ grid world with three rectangular regions $A$, $B$, $C$ enclosed by colored borders. The state space of the robot is its 2D position $s=(x,y)$, which takes 225 discrete combinations. The robot has an action space $A = \{Up, Down, Left, Right\}$. The robot's transition model entails that it follows a given action with probability 0.7, or randomly choose the other three actions each with probability 0.1. The robot has full state observability but does not have knowledge about its transition model. The goal is for the robot to interact with the environment by taking sequences of actions and observing the resultant states, and in the end learn a policy that when followed satisfy the STL specification 

\begin{equation}
\label{eq:7}
\phi = \Box_{[0,\infty)}(\diamondsuit_{[0,40)}\psi_A \wedge \diamondsuit_{[0,40)}\psi_B \wedge \diamondsuit_{[0,40)}\psi_C),
\end{equation}

\noindent where

\begin{equation}
\begin{split}
\psi_A &= (x > 3) \wedge (x < 9) \wedge (y > 10) \wedge (y < 14),\\
\psi_B &= (x > 1) \wedge (x < 5) \wedge (y > 1) \wedge (y < 5),\\
\psi_C &= (x > 9) \wedge (x < 13) \wedge (y > 1) \wedge (y < 7).
\end{split}
\end{equation}

\begin{figure}[tbh]
\label{fig:1}
\begin{center}
\includegraphics[width=0.9\linewidth]{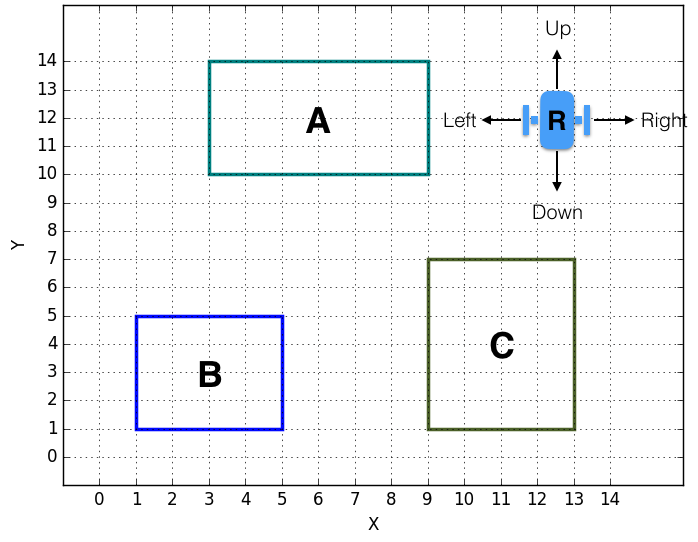}
\caption{A $15 \times 15$ grid world simulation environment. $A,B,C$ are three regions the robot can visit. The robot can choose to move in the four directions shown in the figure. The probability of moving in the desired direction is 0.7 and the probability of moving in any of the three undesired directions is 0.1}
\end{center}
\end{figure}

\noindent In English the above specification says "for as long as the robot is running ($\Box_{[0,\infty)}$), enter regions $A$, $B$ and $C$ every 40 time steps". This is a cyclic task with no termination. Three primitive options are constructed $o_i = \langle I_i, \pi_i, \beta_i \rangle$, $i = A,B,C$. Here we let their initiation sets to be the entire state space i.e. $I_i = S$, which means all three options can be initiated anywhere. The termination map is given by

\begin{numcases}{\beta_i(s)=}
    1  & $s = \underset{s \in S}{\arg\max}\left(r(s,\psi_i)\right)$
     \\
    0 & otherwise,
\end{numcases}

\noindent which indicates that each option only terminates when entering a state where the robustness of that state with respective to the corresponding $\psi_i$ is maximum. The last step is to construct the set of temporally combined options. Here we used $O_T=\{o_A,o_B,o_C,o_{AB},o_{AC},o_{BC},o_{ABC}\}$ (the hypen in the subscript is dropped to save space). Note that the order of subscript is the order in which each primitive option is executed. To obtain a reasonable exploration-exploitation ratio, an $\epsilon-greedy$ exploration policy is carried out. The agent follows the greedy policy (exploitation) with probability $1-\epsilon$, and chooses a random option/action with probability $\epsilon$ (exploration). The $\epsilon-greedy$ exploration is implemented both at the options policy $\mu:S \rightarrow O_T$ and flat policy $\pi:S \rightarrow A$ level. It is important that the flat policies converge faster and takes greedy actions at higher probability than the options policy because execution of options depend on the flat policies. This is enforced by decaying the exploration probabilities $\epsilon$ linearly with time for both $\pi(s)$ and $\mu(s)$ ($\epsilon(t) = \epsilon_0 - d\times t$ where $d$ is the rate of decay) while ensuring that $\epsilon_{flat policies}$ decay faster. The exploration probabilities $\epsilon$ have a lower limit of 0.1 which is to preserve some exploration even near convergence. Table~\ref{tab:2} shows the learning parameters used in simulation. Even though the task specified by the STL formula in Equation~(\ref{eq:7}) is persistent without termination, we divide our learning process in episodes of 200 option choices. That is to say that within each episode the robot chooses an option according to the $\epsilon-greedy$ policy and executes the option to termination, and repeat for 200 times. Then the robot is randomly placed at another location and the next learning episode starts. We performed the training process for 1200 episodes on a Mac with 3 GHz processor and 8 GB memory, and the training took 36 minutes 12 seconds to complete. The resulting policies and two sample runs are presented in Figure (2).

\begin{table}[]
\centering
\caption{Parameters used in simulation}
\label{tab:2}
\begin{tabular}{|c|c|c|}
\hline
\textbf{Parameter} & \textbf{Description} & \textbf{Value} \\ \hline
$\gamma_A, \gamma_B, \gamma_C$ & Discount factors for flat policies & 0.9 \\ \hline
$\alpha_A, \alpha_B, \alpha_C$ & Learning rates for flat policies & 0.2\\ \hline
$\epsilon_{A_0},\epsilon_{B_0},\epsilon_{C_0}$ & \begin{tabular}[c]{@{}c@{}}Initial exploration probability for \\ flat policy learning\end{tabular} & 0.8 \\ \hline
$d_A,d_B,d_C$ & \begin{tabular}[c]{@{}c@{}}Linear decay rate for flat policy's\\ exploration probability\end{tabular} & $10^{-6}$ \\ \hline
$\gamma_o$ & Discount factor for options policy & 0.9\\ \hline
$\alpha_o$ & Learning rate for options policy &  0.5\\ \hline
$\epsilon_{o_0}$ & \begin{tabular}[c]{@{}c@{}}Initial exploration probability for options\\ policy learning\end{tabular} & 0.8 \\ \hline
$d_o$ & \begin{tabular}[c]{@{}c@{}}Linear decay rate for options policy's \\ exploration probability\end{tabular} &  $10^{-4}$\\ \hline
\end{tabular}
\end{table}


\begin{figure*}
 \centering
 \begin{multicols}{3}
  \subcaptionbox{Primitive policy $\pi_A$  \label{fig:2a}}{%
  \includegraphics[width=1.15\columnwidth]{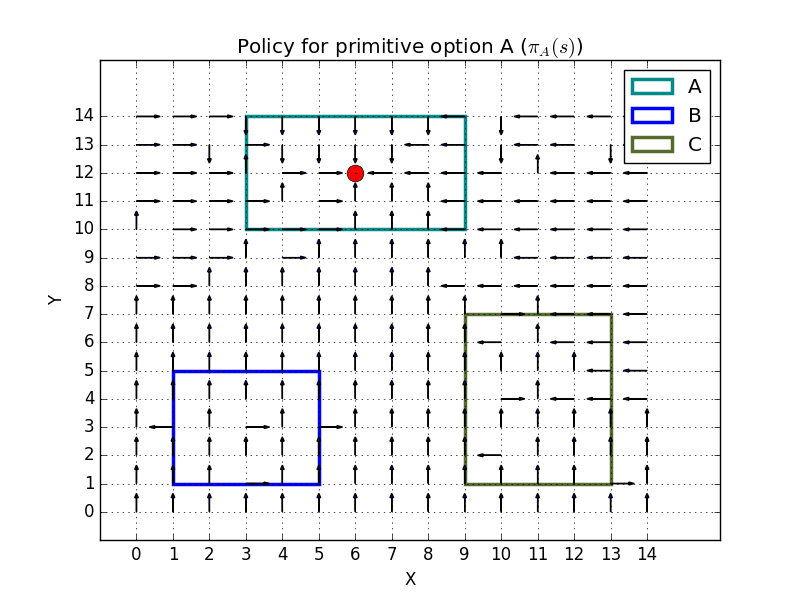}}
  \subcaptionbox{Primitive policy $\pi_B$  \label{fig:2b}}{%
  \includegraphics[width=1.15\columnwidth]{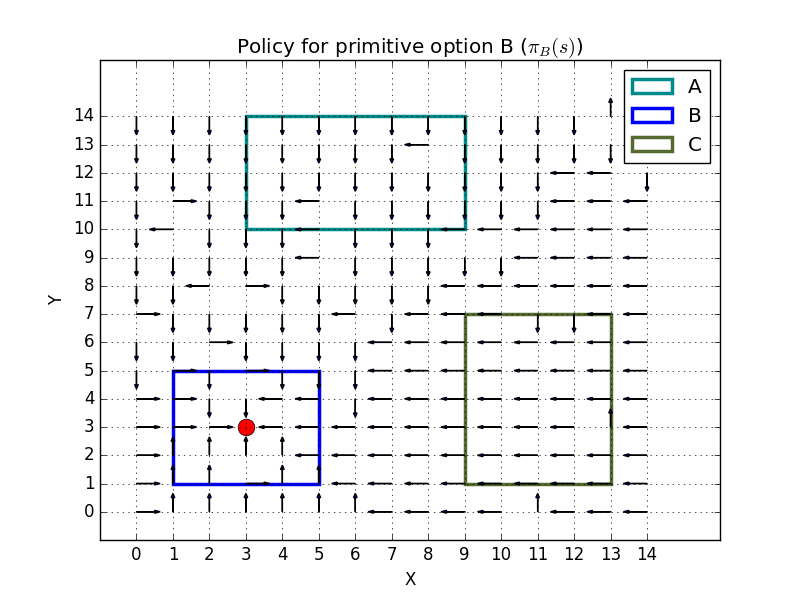} }
  \subcaptionbox{Primitive policy $\pi_C$  \label{fig:2c}}{%
  \includegraphics[width=1.15\columnwidth]{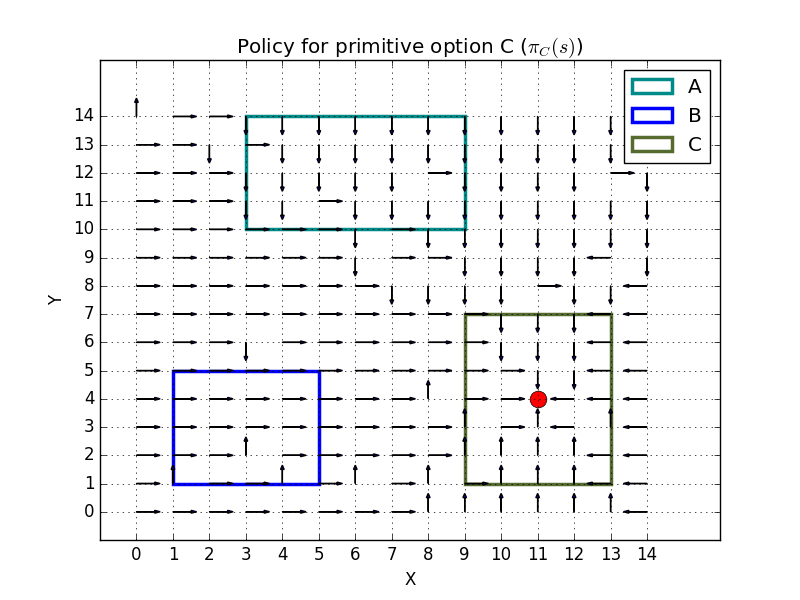} }
  \end{multicols}
  
  \begin{multicols}{3}
   \subcaptionbox{Options policy $\mu$  \label{fig:2d}}{%
  \includegraphics[width=1.15\columnwidth]{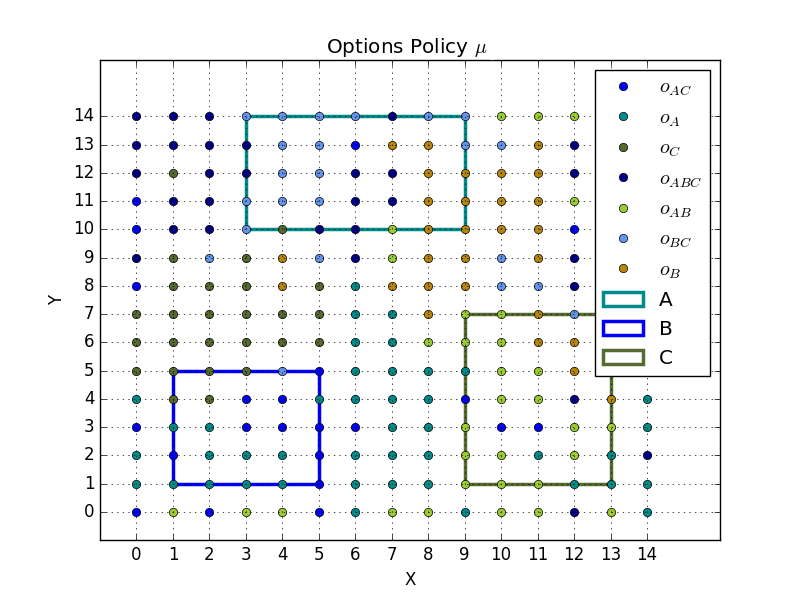} }
 \subcaptionbox{Sample run with \\initial position $s=(14,1)$  \label{fig:2e}}{%
  \includegraphics[width=1.15\columnwidth]{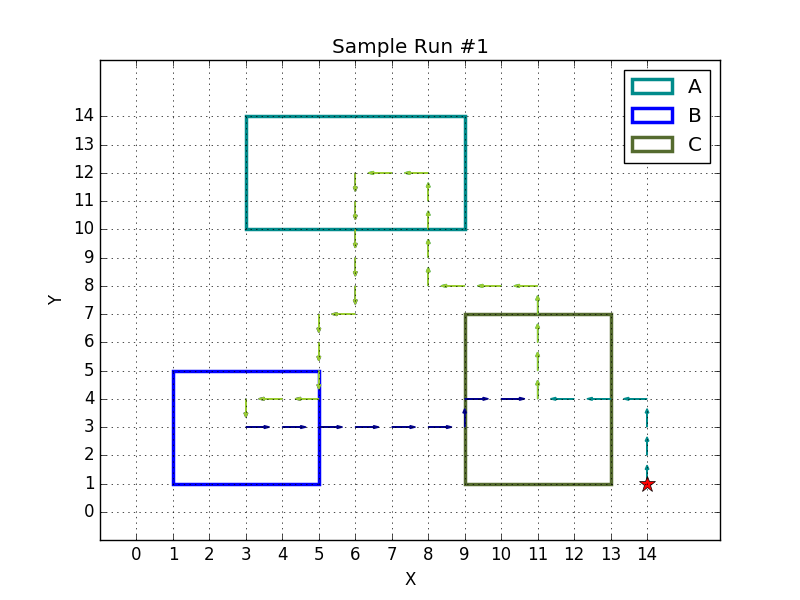} }
   \subcaptionbox{Sample run with \\initial position $s=(7,6)$  \label{fig:2f}}{%
  \includegraphics[width=1.15\columnwidth]{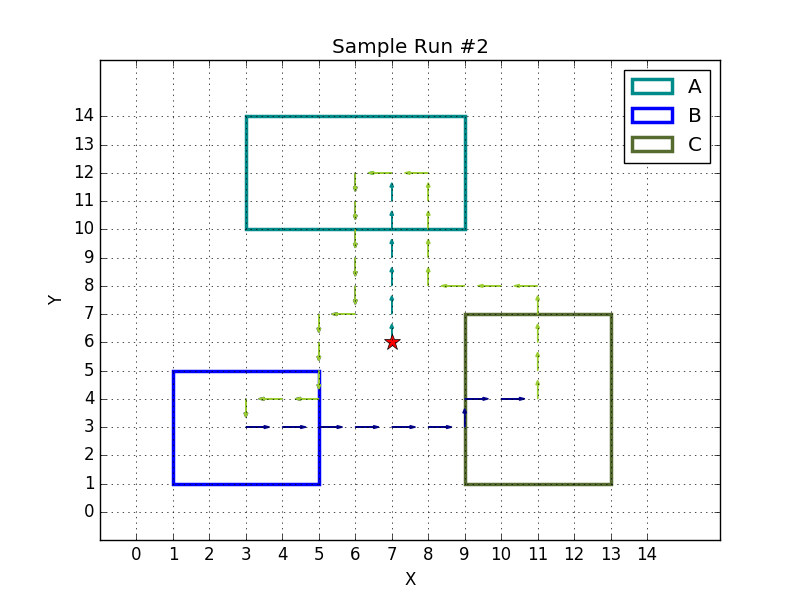} }
  \end{multicols}
  \caption{Learning results for 1200 episodes of training. Subfigures (a), (b), and (c) shows the learned flat policies $\pi: S \rightarrow A$. The red dot in each figure denotes the state of termination defined by the termination map $\beta_i: S \rightarrow [0,1]$. The subfigure (d ) shows the learned options policy $\mu: S \rightarrow O$. Subfigures (e) and (f) illustrate samples run of following the learned policies from two different initial positions shown by the red star}\label{fig:2}
\end{figure*}

Figures (2a), (2b), and (2c) shows the three flat policies $\pi_A$, $\pi_B$, and $\pi_C$ learned by the algorithm. The red dot represents the 
termination state for each option defined by the termination map. In this case the flat policies lead the robot to this state because it is the state of maximum robustness (but 
termination can be any state or set of states defined by the user). Figure (2d) shows the learned options policy $\mu(s)$. This is the policy that 
the robot follows at the highest level. For example $\mu(s)=o_{BC}$ at state $s=(5,11)$, therefore option $o_B$ and $o_C$ will be executed to termination in order. For the
STL formula in Equation (\ref{eq:7}), the desired trajectory as $t \rightarrow \infty$ will be a loop that goes through regions $A$, $B$ and $C$, the action/option taken at any 
other state should lead the agent to this loop along a trajectory that evaluates to the highest robustness degree. Figures (2e) and (2f)
shows two sample runs with different initial positions (indicated by the red star), and the resulting behavior is as expected. The color of the arrows corresponds to the 
color coding of the options in the previous options policy subfigure, and are subject to overlay. It can be observed that although for the 1200 episodes of training neither the flat policies nor the options policy has converged (for example at state $s=(0,14)$ in $\pi_C$), the resulting policies succeed in navigating the robot towards the desired behavior.

As discussed in Section \ref{subsec:3.3}, the quality of the learned policies with respective to maximizing robustness depends on the set of options provided to the algorithm. Figure (3) shows a comparison of cumulative reward per episode between two different sets of temporally combined options. The first is the set $O_{T_1}=O_T$ used in previous simulation. The second set $O_{T_2}=\{o_A,o_B,o_C,o_{AB},o_{BA},o_{AC},o_{CA},o_{BC},o_{CB},o_{ABC},$ $o_{ACB},o_{BAC},o_{BCA},o_{CAB},o_{CBA}\}$ takes into account the permutation of primitive options. Results show that using options set $O_{T2}$ achieves an average of 34.8\% higher cumulative reward per episode compared to using $O_{T2}$ (negative reward values are due to $\epsilon-greedy$ random exploration when following learned policies). However the time used to train the agent for the same 1200 episodes is 43 minutes 51 seconds for $O_{T2}$ compared to 36 minutes 12 seconds for $O_{T1}$. In a way this allows the user to leverage a tradeoff between computational resource and optimality by deciding on the number and complexity of the options provided to the framework. 

\begin{figure}[tbh]
\label{fig:3}
\begin{center}
\includegraphics[width=\linewidth]{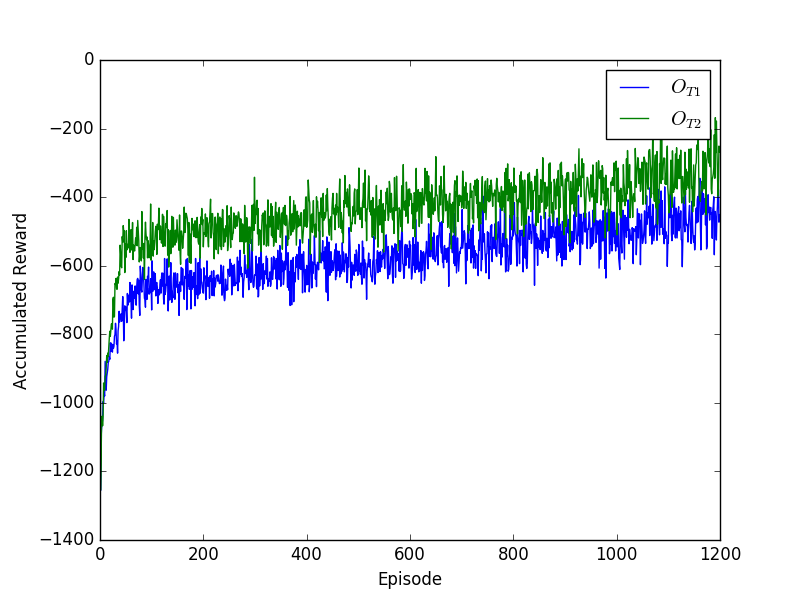}
\caption{Comparison of cumulative reward per episode for two sets of temporally combined options}
\end{center}
\end{figure}


\section{CONCLUSION}
\label{sec:5}

In this paper we have developed a reinforcement learning algorithm that takes in an STL formula as task specification, and learns a hierarchy of policies that maximizes the expected sum of discounted robustness degree with hierarchical optimality. We have taken advantage of the options framework to provide to the learning agent a set of temporally extended actions (options), and the "correctness" of choosing an option at a state is evaluated by calculating the robustness degree of the resulting trajectory against the given STL formula. This naturally becomes the one step immediate reward in the reinforcement learning architecture and thus takes away the burden of manually designing a reward structure. We have shown in simulation that the proposed algorithm learns an options policy and the dependent flat policies that guide the agent to satisfy the task specification with a relatively low number of training steps. The temporal and state abstraction provided by options and STL respectively decomposes a complicated task into a hierarchy of simpler subtasks, and thus modularizing the learning process and increasing the learning efficiency. Moreover, the policies learned for the subtasks can be reused for learning a different high level task and therefore knowledge transfer is enabled. In future work we will look at applying the proposed algorithm to more realistic problems and extending from discrete state and action spaces to continuous ones.




\bibliographystyle{IEEEtran}
\bibliography{cdc2016}

\end{document}